\documentclass[10pt,twocolumn,letterpaper]{article}

\usepackage{cvpr}              

\usepackage{graphicx}
\usepackage{amsmath}
\usepackage{amssymb}
\usepackage{booktabs}

\usepackage[pagebackref,breaklinks,colorlinks]{hyperref}
\usepackage{lipsum}

\newcommand\blfootnote[1]{%
  \begingroup
  \renewcommand\thefootnote{}\footnote{#1}%
  \addtocounter{footnote}{-1}%
  \endgroup
}

\usepackage[capitalize]{cleveref}
\crefname{section}{Sec.}{Secs.}
\Crefname{section}{Section}{Sections}
\Crefname{table}{Table}{Tables}
\crefname{table}{Tab.}{Tabs.}

\def\cvprPaperID{20} 
\def\confName{CVPR}
\def\confYear{2023}

\begin{document}

\title{PWR-Align: Leveraging Part-Whole Relationships for Part-wise Rigid Point Cloud Registration in Mixed Reality Applications}

\author{Manorama Jha\\
GridRaster Inc.\\
{\tt\small manorama@gridraster.com}
\and
Bhaskar Banerjee\\
GridRaster Inc.\\
{\tt\small bhaskar@gridraster.com}
}
\maketitle

\begin{abstract}
We present an efficient and robust point cloud registration (PCR) workflow for part-wise rigid point cloud alignment using the Microsoft HoloLens 2. Point Cloud Registration (PCR) is an important problem in Augmented and Mixed Reality use cases, and we present a study for a special class of non-rigid transformations. Many commonly encountered objects are composed of rigid parts that move relative to one another about joints resulting in non-rigid deformation of the whole object such as, robots with manipulators, and machines with hinges. The workflow presented allows us to register the point cloud with various configurations of the point cloud.   
\end{abstract}

\vspace{-6pt}
\section{Introduction}
\label{sec:intro}
Mixed Reality (MR) \cite{speicher2019mixed} is used to create immersive experiences by blending digital entities with the real-world environment. Point Cloud Registration (PCR) \cite{pomerleau2015review}, also called alignment, is the task of finding a spatial transformation (rotation and translation) which when applied to a source point cloud results in point-wise superposition with a destination point cloud of the same object and/or scene. One of the primary use cases of PCR in MR is to overlay a virtual reference model of an object on top of the corresponding real object. For example, in manufacturing processes a virtual reference model of a machine part can be overlaid on a damaged instance for inspection \cite{munoz2019mixed}, repair \cite{abate2013mixed,borsci2015empirical} and training \cite{mueller2003marvel,kirkley2005creating,wang2004mixed}. 
Real world applications of PCR face several challenges like lack of order or structure \cite{lawin2017deep,maturana2015voxnet,roynard2018classification,ben20183dmfv,thomas2019kpconv}, sparsity \cite{choy20194d}, noise and outliers, \cite{bai2021pointdsc,pais20203dregnet,holz2015registration,schnabel2007efficient} and partial overlap \cite{thomas2019kpconv,xu2021omnet, attaiki2021dpfm, rodola2017partial} of the source and target point clouds. \blfootnote{Accepted for presentation at the Women in Computer Vision (WiCV) workshop at CVPR-2023.}

\subsection{Background and Related Work}
\label{sec:related_work}
 Traditionally, PCR involves two steps - a) feature extraction \cite{weinmann2017geometric,qi2017pointnet,thomas2019kpconv} from the source and target point clouds and b) feature matching \cite{holz2015registration,schnabel2007efficient,arun1987least,besl1992method,zhang1994iterative} to detect corresponding pairs of points in the two clouds. The correspondence matches are then used to derive a transformation function that maps the source to the target such that the corresponding points overlap. For rigid objects, this transformation function is an affine transformation consisting of rotation and translation. For non-rigid objects, this transformation function is more complex and it involves warping the surface of the source point cloud on to the target. Non-rigid Iterative Closest Point \cite{li2008global,li20214dcomplete} is a state-of-the-art algorithm that applies a locally affine regularization that assigns an affine transformation to each vertex and minimizes the difference in the transformation of neighboring vertices. With this regularization, the optimal deformation for fixed correspondences and fixed stiffness can be determined precisely and efficiently. The algorithm loops over a series of decreasing stiffness weights that results in incremental deformation of the template surface towards the target.

\begin{figure}[!t]
    \centering
    \includegraphics[width=0.8\linewidth]{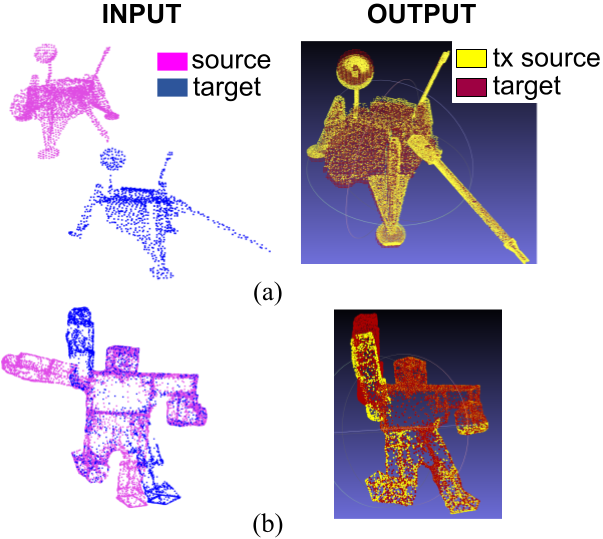}
    \caption{Sample registration results for (a) NASA Viking Lander \cite{viking2016nasa} and (b) Cubebot \cite{cubebot2016sketchfab} models. The target point cloud is scanned using Microsoft Hololens 2 MR headset under real-world conditions. Our workflow achieves robust and accurate performance even in the presence of sparse, noisy, and partial data.}
    \label{fig:results}
\end{figure}


The scope of this paper is limited to finding a non-rigid transformation using the set of predicted correspondences. In the general case, this involves finding a transformation for each point on the source mesh and the computational requirement can blow up for large meshes. Additional challenges are posed by noise, holes and missing parts in real world 3D scans. Computational efficiency also becomes crucial in MR applications to produce a lag-free immersive experience. 

\subsection{Contributions}
\label{sec:contributions}
In this paper, we study PCR for a special class of non-rigid transformations that happen in objects composed of rigid parts that can move relative to one another (e.g., about a hinge). These are commonly encountered in enterprise MR applications, e.g. machine parts, doors, wheels, antennas, etc. For such objects, the non-rigid adjustment step after rigid fitting requires finding a rigid transformation for each part that can move relative to one another. We begin by constructing a graph to represent the part-whole relationships in the source object. We present a novel approach to derive a rigid transformation for each part, leveraging this graph to -- a) make sure that the object model does not disintegrate after applying part-wise registration; b) search for a given part in the correct part of the target point cloud; and c) handle outliers in the correspondence map.\\

\section{Proposed Method - PWR-Align}
\label{sec:proposed_method}
In this section, we present PWR-Align, our proposed workflow that leverages \underline{P}art-\underline{W}hole \underline{R}elationships for \underline{Align}ment. We start with \textbf{constructing a bidirectional graph} of the source point cloud (from a CAD model or digital twin). Each node of this graph represents one of the rigid parts. Each pair of adjacent parts is connected with an edge. The next step is \textbf{finding corresponding points} in the source and target point clouds for which we use the Lepard \cite{li2022lepard} non-rigid correspondence matching algorithm. The third step is the \textbf{whole body rigid fitting}, where we derive the rigid body transformation for the whole source point cloud. This aligns the source and target point clouds with possible misalignments of movable parts. The final step involves \textbf{part-wise tuning} where each part is tuned in two steps - first using RANSAC \cite{holz2015registration,schnabel2007efficient} and then using the Iterative Closest Point (ICP) \cite{wang2017survey} algorithms. The output of this step is a rotation and translation matrix for the part.\\

\noindent To prevent the object model from disintegrating upon part-wise transformation we take the following steps:
   \begin{enumerate}
     \item We adjust the parts in decreasing order of sizes.
     \item When using RANSAC, we add extra correspondences in the form of points that are at the junction of the current part and its larger neighbor that must remain at the same position where they were before the adjustment.
     \item If application of a transformation to a part causes a joint to be broken, we skip the adjustment.
     \item If the number of detected correspondences is too few for a part, we skip the RANSAC step for that part.
    \end{enumerate}
To make sure that a part is searched for in the correct part of the target point cloud, we take the following measures:
   \begin{enumerate}
     \item We perform whole body rigid fitting as the first step to roughly align the two point clouds.
     \item We search for a given part only within a Region-of-Interest (RoI) of the target point cloud. The RoI is calculated as the smallest axis-aligned bounding box that contains the part in the source point cloud after whole body rigid fitting and all the corresponding points in the target point cloud for the feature points of the part.
    \end{enumerate}
Finally, in order to improve robustness against outliers, we take the following measures:
    \begin{enumerate}
     \item KPConv applies a uniform downsampling to prevent the compute load (quadratic in the number of samples) from blowing up. If the density of a point cloud is low, this leads to serious loss in detail and significant degradation in performance. We make sure there are a minimum of $50$ points in each part of the input point clouds. 
     \item We perform ICP after RANSAC to correct for the effect of spurious correspondences.
    \end{enumerate}
\section{Results and Discussion}
\label{sec:results}

We evaluated our workflow on data collected using Microsoft Hololens 2 MR headset \cite{ungureanu2020hololens} under real world conditions. Fig.\ref{fig:results} shows two sample outcomes for NASA Viking Lander \cite{viking2016nasa} and Cubebot \cite{cubebot2016sketchfab}. We observe that our workflow is able to perform accurately even in the presence of noise, sparsity and missing parts in the target point cloud scan. Performance degrades gracefully as sparsity increases. Fig.\ref{fig:results}~(b) presents an extreme case where the target point cloud is a single-view scan. As expected, we can observe some misalignment near the deformed hand of the Cubebot. Please refer to the full thesis \cite{jha2022point} for a detailed discussion of the experiments and results.
\section{Conclusion}
\label{sec:conclusion}
In this paper, we study the problem of point cloud registration for a special class of non-rigid deformations that are encountered in objects composed of rigid parts that move about joints. We propose an efficient and robust workflow called PWR-Align that leverages part-whole relationships. We evaluate our method on real world MR data and demonstrate its robustness and viability as a faster alternative to non-rigid ICP. 
{\small
\bibliographystyle{unsrt}
\bibliography{egbib}
}

\end{document}